\documentclass[conference]{IEEEtran}
\usepackage{graphicx,times,amsmath,amssymb}
\usepackage{epsfig,algorithm,algorithmic,url}

\urldef{\mailsa}\path|{tmaszczyk, wduch}@is.umk.pl|

\newcommand\vm{\vec{m}}
\newcommand\vx{\vec{x}}
\newcommand\vA{\vec{A}}
\newcommand\vX{\vec{X}}
\newcommand\vw{\vec{w}}
\newcommand\vz{\vec{z}}
\newcommand\be{\vspace*{-1pt}\begin{equation}} 
\newcommand\ee{\end{equation}\vspace*{-1pt}}
\newcommand\bea{\vspace*{1pt}\begin{eqnarray}}
\newcommand\eea{\end{eqnarray}\vspace*{1pt}}

\begin{document}

\title{\ \\ \LARGE\bf Support Feature Machines:\\ 
Support Vectors are not enough. 
\thanks{Tomasz Maszczyk \and W\l{}odzis\l{}aw Duch\\
Department of Informatics, Nicolaus Copernicus University\\
Grudzi\c{a}dzka 5, 87-100 Toru\'n, Poland\\
\mailsa\\
\url{http://www.is.umk.pl}}} 

\author{Tomasz Maszczyk and W\l{}odzis\l{}aw Duch} 

\maketitle

\begin{abstract}
Support Vector Machines (SVMs) with various kernels have played dominant role in machine learning for many years, finding numerous applications. Although they have many attractive features interpretation of their solutions is quite difficult, the use of a single kernel type may not be appropriate in all areas of the input space, convergence problems for some kernels are not uncommon, the standard quadratic programming solution has $O(m^3)$ time and $O(m^2)$ space complexity for $m$ training patterns. Kernel methods work because they implicitly provide new, useful features. 
Such features, derived from various kernels and other vector transformations, may be used directly in any machine learning algorithm, facilitating multiresolution, heterogeneous models of data. 
Therefore Support Feature Machines (SFM) based on linear models in the extended feature spaces, enabling control over selection of support features, give at least as good results as any kernel-based SVMs, removing all problems related to interpretation, scaling and convergence. This is demonstrated for a number of benchmark datasets analyzed with linear discrimination, SVM, decision trees and nearest neighbor methods. 

\end{abstract}

\section{Introduction}
The most popular type of SVM algorithm with localized (usually Gaussian) kernels \cite{Scholkopf01} suffers from the curse of dimensionality \cite{bengio05}. This is due to the fact that such algorithms rely on assumption of uniform resolution and local similarity between data samples. To obtain accurate solution often a large number of training examples used as support vectors is required. This leads to high cost of computations and complex models that do not generalize well. Much effort has been devoted to improvements of the scaling \cite{CVM:Tsang06,SVMPrimal:Chapelle07}, reducing the number of support vectors, 
introducing relevance vectors \cite{RVM:Tipping01}, and improving (learning) multiple kernel design 
\cite{Shogun:Sonnenburg06}. All these developments are impressive, but there is still room for simpler, more direct and comprehensible approaches. 

Kernel methods work because they implicitly provide new, useful features 
$z_i(\vx)=k(\vx,\vx_i)$ 
constructed around support vectors $\vx_i$, a subset of input vectors relevant to the training objective. Prediction is supported by new features, and these features do not need to be local or connected to single reference vectors. Therefore this approach is called here "Support Feature Machine", rather than vector machine. It is related to the idea of "learning from the successes of others", implemented in our Universal Learning Machines \cite{ULM09}, where data models created by different algorithms are analyzed to discover the most useful transformations (prototypes, linear combinations, branches in decision trees), that are then added to the pool of expanded features. 
In the final feature space almost all machine learning algorithms perform at the same level. The choice of the algorithm becomes then a matter of preference, but various algorithms are still needed to discover useful ``knowledge granules'' in data. For example, local features used by the nearest-neighbor methods may be very useful, and they are provided by localized kernels. At the same time various projections may also be very useful. 

This approach is also a step towards meta-learning, general framework for creating optimal adaptive systems on demand for a given problem \cite{DuchMeta02,GrabczewskiJ08}. The type of solution offered by a given data model obtained by SVM with a specific kernel may not be appropriate for the particular data. Each data model defines a hypotheses space, that is a set of functions that this model may easily learn. Linear methods work best when decision border is flat, but they are obviously not suitable for spherical distributions of data, requiring $O(n^2)$ parameters to approximately cover each spherical distribution in $n$ dimensions, while an expansion in radial functions requires only $O(n)$ parameters. For some problems (for example, high-dimensional parity and similar functions), neither linear nor radial decision borders are sufficient \cite{DuchKsep06}. An optimal solution may only be found if a model based on quasi-periodic non-linear transformations is defined \cite{ULM09}. 

Support Feature Machines introduced here are specific generalization of SVMs. In the second section standard approach to the SVM is described and linked to evaluation of similarity to support vectors in the space enhanced by $z_i(\vx)=k(\vx,\vx_i)$ kernel features. Linear models defined in the enhanced space are equivalent to kernel-based SVMs. In particular, one can use linear SVM to find discriminant in the enhanced space, preserving the wide margins. For special problems other techniques may be more appropriate \cite{LDA:Tebbens07}. 
With explicit representation of features interpretation of discriminant function is straightforward. 
Kernels with various parameters may be used, including degree of localization, and the resulting discriminant may select global features combined with local features that handle exceptions. 
New features based on non-local projection and partially localized projections are introduced and added to the pool of all features. Original input features may also be added to the support features, although they are rarely of comparable importance. This guarantees that the simplest solutions to easy problems are not overlooked. Support Features Machines are simply linear discriminant functions defined in such enhanced spaces. 
In section 4 SFMs are tested in a number of benchmark calculations, and usefulness of additional features in approaches as diverse as decision trees and nearest neighbor methods is demonstrated. In all cases improvements over the single-kernel SVM results are obtained.
Brief discussion of further research directions concludes this paper.

\section{Kernels and Support Vector Machines}
\subsection{Standard SVM formulation}

Since the seminal paper of Boser, Guyon and Vapnik in 1992 \cite{SVM:Boser92} Support Vector Machines quickly became the most popular method of classification and regression, finding numerous other applications \cite{Scholkopf01,Scholkopf98,Diederich08}. 
In case of binary classification problems SVM algorithm minimizes average errors (or risk) over the set of data pairs $\langle x_i,y_i \rangle$. 
Depending on the choice of kernels and optimization of their parameters SVM can produce flexible nonlinear data models that, thanks to the optimization of classification margin, offer good generalization. This means that the minimum distance between the training vectors $\vx_i$ and the hyperplane $\vw$ should be maximized:
\be
\max_{\vw,b} \min{\|\vx-\vx_i\| \ : \ \vw\cdot\vx+b=0, \ i=1,\dots,m}
\ee
The $\vw$ and $b$ can be rescaled in such a way that the point closest to the hyperplane $\vw\cdot\vx+b = 0$, lies on one of the parallel hyperplanes defining the margin $\vw\cdot\vx+b = \pm 1$. 
This leads to the requirement that 
\be
\forall_{\vx_i} \ y_i[\vw\cdot\vx_i+b]\geq 1
\ee
The width of the margin is equal to $2/\|w\|$.
The problem can be restated as maximization of margins:
\be \label{eq:tau}
\min_{\vw,b} \ \tau(\vw)=\frac{1}{2}\|\vw\|^2
\ee
with constraints that guarantee correct classification:
\be
y_i[\vw\cdot\vx_i+b]\geq 1 \ \ \ \ i=1,\dots,m
\ee
Constraint optimization problems are solved by defining Lagrangian:
\be
L(\vw,b,\alpha)=\frac{1}{2}\|\vw\|^2-\sum_{i=1}^m\alpha_i(y_i[\vx_i\cdot\vw+b]-1)
\ee
where $\alpha_i > 0$ are Lagrange multipliers. 
Its minimization over $b$ and $\vw$ leads to two conditions:
\be
\sum_{i=1}^m\alpha_i y_i = 0, \ \ \ \ \vw=\sum_{i=1}^m\alpha_i y_i\vx_i
\ee
The vector $\vw$ that defines the hyperplane is expressed as a combination of the training vectors, each component $\vw[j]$ is a combination of $j$ feature values for all vectors $\vx_i[j]$. 
According to the Karush-Kuhn-Thucker conditions:
\be
\alpha_i(y_i[\vx_i\cdot\vw+b]-1)=0, \ \ \ \ i=1,\dots,m
\ee
For $\alpha_i \neq 0$ vectors must lie on one of the margin hyperplanes 
$y_i[\vx_i\cdot\vw + b] = 1$; these vectors ``support'' the hyperplane $\vw$ that defines the solution of the optimization problem. 
Although the minimization may be performed in the primal form \cite{SVMPrimal:Chapelle07} 
the quadratic optimization problem is frequently redefined in a bit simpler dual form:
\be
\max_\alpha \ \vw(\alpha)=\sum_{i=1}^m\alpha_i-\frac{1}{2}
\sum_{i,j=1}^m \alpha_i \alpha_j y_i y_j \vx_i\vx_j
\ee
with constraints:
\be
\alpha_i\geq 0 \ \ i=1,\dots,m \ \ \ \ \sum_{i=1}^m\alpha_i y_i=0
\ee
The discriminant function takes the form:
\be
g(x)=\hbox{sgn}\left(\sum_{i=1}^m\alpha_iy_i \vx\cdot\vx_i+b\right)
\ee
Now it is easy to replace dot product $\vx\cdot\vx_i$ by a kernel function $k(\vx,\vx')=\phi(\vx)\cdot\phi(\vx')$ where $\phi(\vx)$ represents an implicit transformation (because only the kernel functions is used) of the original vectors to a new space. 
Usually the Cover theorem \cite{Cover65} is invoked to justify mapping to higher-dimensional spaces. 
However, for any $\phi(\vx)$ vector the part orthogonal to the space spanned by $\phi(\vx_i)$ does not contribute to $\phi(\vx)\cdot\phi(\vx')$ products, so it is sufficient to express $\phi(\vx)$ and $\vw$  as a combination of $\phi(\vx_i)$ vectors. The dimensionality $n$ of the input vectors is frequently lower than the number of training patterns $n<m$, and then $\phi(\vx)$ represents mapping into higher $m$-dimensional space. In the microarray data and some other problems the reverse situation is true: dimensionality is much higher than the number of patterns for training. 

The discriminant function in the $\phi()$ space is:
\be
g(\vx)=\hbox{sgn}\left(\sum_{i=1}^m\alpha_i y_i k(\vx,\vx_i)+b\right)
\ee
If the kernel function is linear the $\phi()$ space is simply the original space and the contributions to the discriminant function are based on the cosine distances to the reference vectors $\vx_i$ from the $y_i$ class. Thus the original features $\vx[j], j=1..n$ are replaced by new features 
$z_i(\vx)=k(\vx,\vx_i)$ that evaluate how close (or how similar) the vector is from the training vectors. Incorporating signs in the coefficient vector $A_i=\alpha_i y_i$ discriminant functions is: 
\be
g(\vx) = \hbox{sgn}\left(\sum_{i=1}^m\alpha_i y_i z_i(\vx))+b\right)=\hbox{sgn}\left(\vA\cdot\vz(\vx))+b\right)
\ee
With the proper choice of non-zero $\alpha$ coefficients this functions is a distance measure from support vectors that are at the margins. In non-separable case instead of using cosine distance measures it is better to use localized similarity measures, for example by scaling  the distance with Gaussian functions; this leads to one of the most useful kernels:
\be
k_G(\vx,\vx')=\exp(-\beta\|x-x'\|^2)
\ee

Many specialized kernels for structured problems, trees, sequences and other types of data may be devised, measuring various aspects of similarity, important for a given task. 
Kernel-based methods use similarity in a special way in combination with linear discrimination, but similarity matrices may also be used in many other ways \cite{SBMneural00,DuchSBM00}.

\section{Support Feature Machines}

For each vector $\vx$ we have not only $n$ input features but also $m$ kernel features $z_i(\vx)=k(\vx,\vx_i)$ defined for each training vector. Taking the Gaussian kernel $k_G(\vx,\vx')$ and fixing the value of discriminant $g(\vx)=$constant is equivalent to taking a weighted sum of Gaussians centered at some support vectors that are near the border (for large dispersion all vectors may contribute, but will not influence decision borders). Because contours of discriminant function in the kernel space are approximately constant when $\vx$ moves along the non-linear decision border in the input space, they lie on the hyperplane in the kernel space. Therefore in the space of kernel features linear discriminant methods may be applied directly, without the SVM machinery. 
This will be demonstrated in computational experiments by comparing the results of SVM with Gaussian kernel solved by quadratic programming with direct linear solutions in the kernel-based feature space.

In some cases the use of kernel features is an overkill, as separation may be achieved using original features that are not present in the kernel space. Suppose that data for each class have Gaussian distributions (which is frequently the case), then the best separation direction is simply equal to the difference of sample means $\vw=\vm_1-\vm_2$. Adding projection on this direction as a new feature $r(\vx)=\vw\cdot\vx$ will allow linear discrimination to find simple solution. Note, however, that minimization of $\tau(\vw)$ (Eq. \ref{eq:tau}) to achieve large margin is not going to find simple binary solution, the preference is rather to find more complex solutions with many small coefficients $w_i$. There are other linear discriminant methods that may be used instead \cite{Webb02}, but we shall not pursue this problem further here.

The SFM approach is based on generation of new ``support features'' (SFs) using various kernels, random linear projections, and restricted projections, followed by feature selection and linear discrimination. We shall also consider other machine learning algorithms in the space enhanced by support features. 
In this paper only restricted version of this approach is implemented (see Algorithm \ref{alg:SFM}) using three types of features described below. 

Features of the first type are made using projections on $N$ randomly generated directions in the original $n$-dimensional input space. These directions may be improved in a systematic way, for example by adding directions connecting the means of class-dependent clusters, but this option has not been explored. 
A sufficient number of random directions increases dimensionality and, according to the Cover theorem \cite{Cover65}, allows for easier separation of the data. There is a large literature on random projections and some successes in random initialization of input layers with linear discrimination for the output layer \cite{ELM06}. 

The second type of features is based on restricted random projections, as used in our almost Random Projection Machine (aRPM) approach \cite{aRPM09}. Projections on a random direction 
$z_i(\vx) = \vw_i\cdot\vx$ may not be very useful as a whole, but in some $z_i$ range of values there may be a sufficient large pure cluster of projected patterns. For example, in case of parity problems \cite{DuchKsep06,GrochowskiD07} projections always have strong overlaps of class-conditional probability distributions, but projections on $[1,1..1]$ direction show pure localized clusters with fixed number of 1's. Clusters containing training patterns from class $C$ may be separated from other patterns projected on $z_i$ dimension, defining window-like functions $h_i(\vx)=H(z_i(\vx);C)$. For example, bicentral functions 
\cite{DuchJankowskiNCS99} 
equal to a difference of two logistic functions, provide a soft trapezoidal windows 
$H(z_i(\vx);C)=\sigma(z_i-a)-\sigma(z_i+b)$. Below only a simple $[a,b]$ intervals have been used. 
This creates binary features $h_i(\vx)\in\{0,1\}$, based on linear projection restricted to a slice of the input space perpendicular to the $z_i$ dimension. We have also used here directions from the Quality of Projected Clusters (QPC) projection pursuit index 
\cite{GrochowskiD08a} that allows for tuning these directions to increase cluster sizes. 

The third type are features based on kernels. While many kernels may be mixed together, including the same kernels with different parameters, in the initial implementation only Gaussian kernels with a fixed dispersion $\beta$ are taken for each training vector (potential support vector) $k_i(\vx)=\exp(-\beta\sum|\vx_i-\vx|^2)$. Training vectors that are far from decision borders may of course be removed in many different ways, but again in this initial implementation of the SFM approach this has not been considered. 

Generation of features is linear in the number of training patterns $m$, but for large $m$ it should be reduced using simple filters \cite{Duchfilter06}. Recently we have developed a new library for feature ranking, selection and redundancy removal \cite{Infosel:Kachel10} that is well suited for this purpose. 
Here only the simplest version based on mutual information filter is used. Local kernel features have values close to zero except around their support vectors. Therefore their usefulness should be limited to the neighborhood $O(\vx_i)$ in which $G_i(\vx)>\epsilon)$ (this has been set to $\epsilon=0.001$). Similarly for restricted projections the neighborhood is restricted to vectors that fall into the interval $[a,b]$ with single-class patterns. Strongly localized features used in the Naive Bayes algorithm will lead to a majority voting rule, therefore this algorithm has not been used here. 

To accept a new feature $f$ of the $z, h, k$ type after it has been generated three conditions should be met: 

\begin{enumerate}
\item neighborhoods should not be too small, local features should cover at least $\eta$ vectors;
\item in local neighborhood $MI(f(\vx),C)>\alpha$, mutual information of feature $f(\vx)$ should not be too small; 
\item maximum probability $\max_C p(C|f(\vx))>\delta$ selects those features $f(\vx)$ that discriminate between classes. 
\end{enumerate}

Number of vectors in the neighborhood $\eta$ has been arbitrarily set to $\eta=10$, although in some applications with very few training vectors lower values could be considered. 
Unrestricted projections cover all data and cannot have $p(C|z(\vx))=1$ for all vectors, so only mutual information is used to select them. 
Parameters $\alpha$ and $\delta$ are set to leave sufficient number of useful features based on kernels supported by vectors near the decision border, or restricted projections that also fall close to the border. These parameters have been fixed to leave 0.3$m$ vectors for each dataset. Their influence on the selection of support vectors for kernels (and thus selection of localized features) is shown in Fig.1-3, where two overlapping Gaussian distributions are used. Of course in this case none of these localized kernels will be finally left in the discriminant function, as the projection on the line connecting sample means is the single feature that is sufficient. Small $\alpha\approx 0.005$ and $\delta$ around 0.5 will leave only vectors around decision borders. 

Parameter $\beta$ may be controlled by the user to determine the degree of smoothness. It may also be automatically set in two ways. First, instead of regulating the smoothness of decision borders by the density of kernels with fixed neighborhood size the distance to the nearest vectors from other classes may be used to set it. Second, several fixed values of $\beta$ may be used, with feature ranking taking care of accepting local features at the required resolution. In calculations reported below fixed value of $\beta=2^{-5}$ has been used.

The final vector $\vX$ is thus composed from a number of $\vX=[x_1,.. x_n z_1,.. h_1,.. k_1...]$ features. In SFM linear solution is sought in this space, but in this extended feature space other learning models may find even better solution.

\begin{algorithm}
\caption{Support Feature Machine}
\label{alg:SFM}
\begin{algorithmic}[1]
\REQUIRE Fix the values of $\alpha$, $\beta$, $\delta$ and $\eta$ parameters.
\medskip
\FOR {$i=0$ to $N$}
\STATE Randomly generate new direction $\vw_i \in [0,1]^n$
\STATE Project all $\vx$ on this direction $\vz_i = \vw_i\cdot\vx$ (features $z$)
\STATE Analyze $p(z_i|C)$ distributions to determine if there are pure clusters, 
\IF {the number of vectors in cluster $H_j(z_i;C)$ exceeds $\eta$} 
\STATE accept new binary feature $h_{ij}$
\ENDIF 
\ENDFOR \\
\STATE Create kernel features $k_i(\vx), i=1..m$
\STATE Rank all original and additional features $f_i$ using Mutual Information. 
\STATE Remove features for which $MI(k_i,C)\leq\alpha$. 
\STATE Remove features for which $\max_C p(C|f(\vx))<\delta$. 
\STATE Build linear model on the enhanced feature space. 
\STATE Classify test data mapped into enhanced space. 
\end{algorithmic}
\end{algorithm}

New support features created in this way are based on those transformations of inputs that have been found interesting for some task, and thus have some meaning and interpretation. Support features are not learned, but selected from random projections, or constructed with the help of localized kernel functions, and added if they show interesting correlations with some aspect of the problem being solved. On a more technical level this means that more attention is paid to generation of features rather than to the sophisticated optimization algorithms or new classification methods. The importance of generating new features has already been stressed in our earlier papers 
\cite{ULM09,aRPM09,Meta:Maszczyk10}, but adding kernel features in SFM proved to be essential for improving upon kernel-based SVMs. 
In essence SFM requires construction and selection of new features, followed by simple linear models of learning. Although several parameters may be used to control the process they are either fixed or set in an automatic way. SFM solutions are highly accurate and easy to understand. Neurobiological justification of such approach is presented in the final discussion. 

\begin{figure}[!htb]
\begin{center}
\includegraphics[width=3.35in]{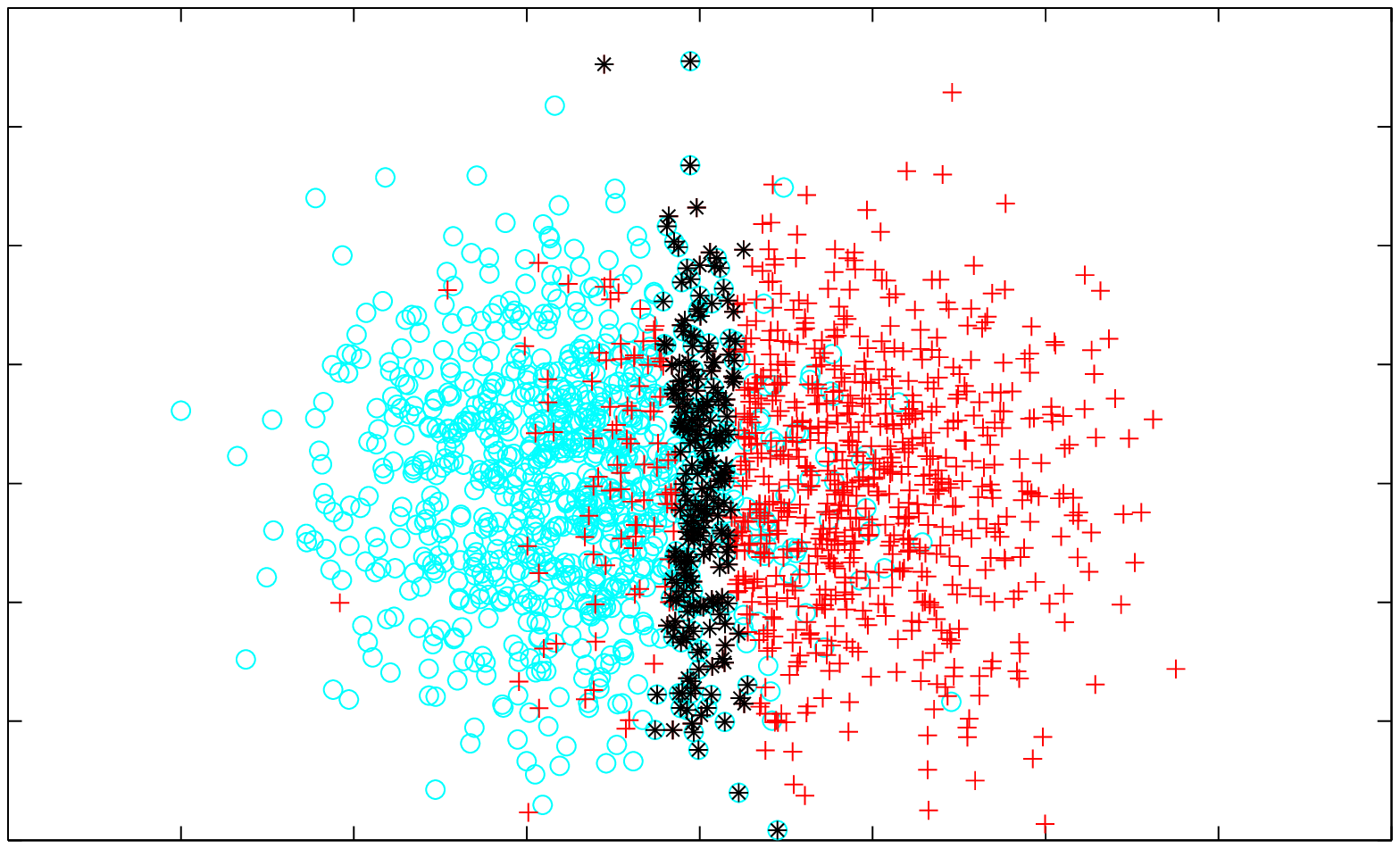}
\includegraphics[width=3.35in]{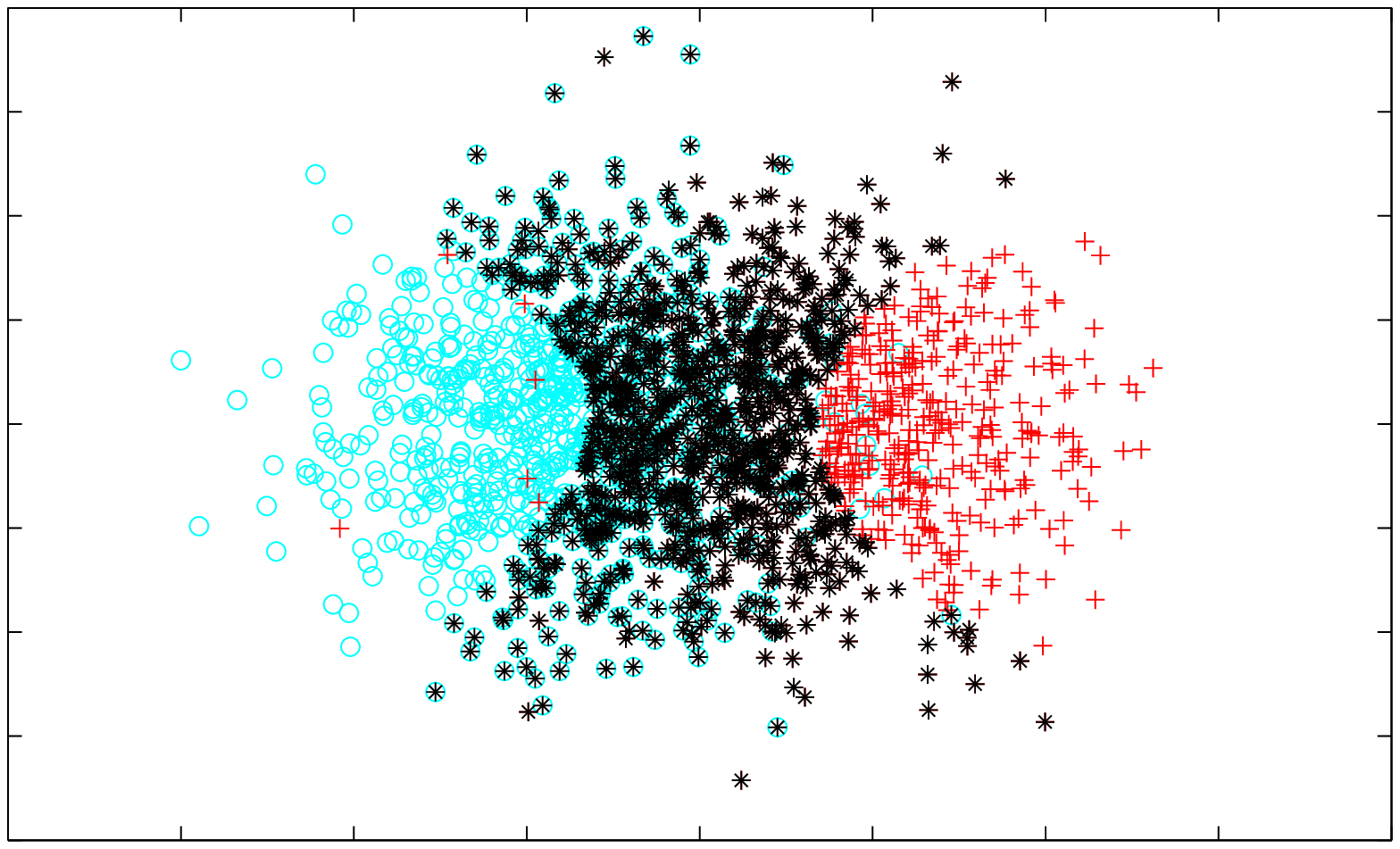}
\includegraphics[width=3.35in]{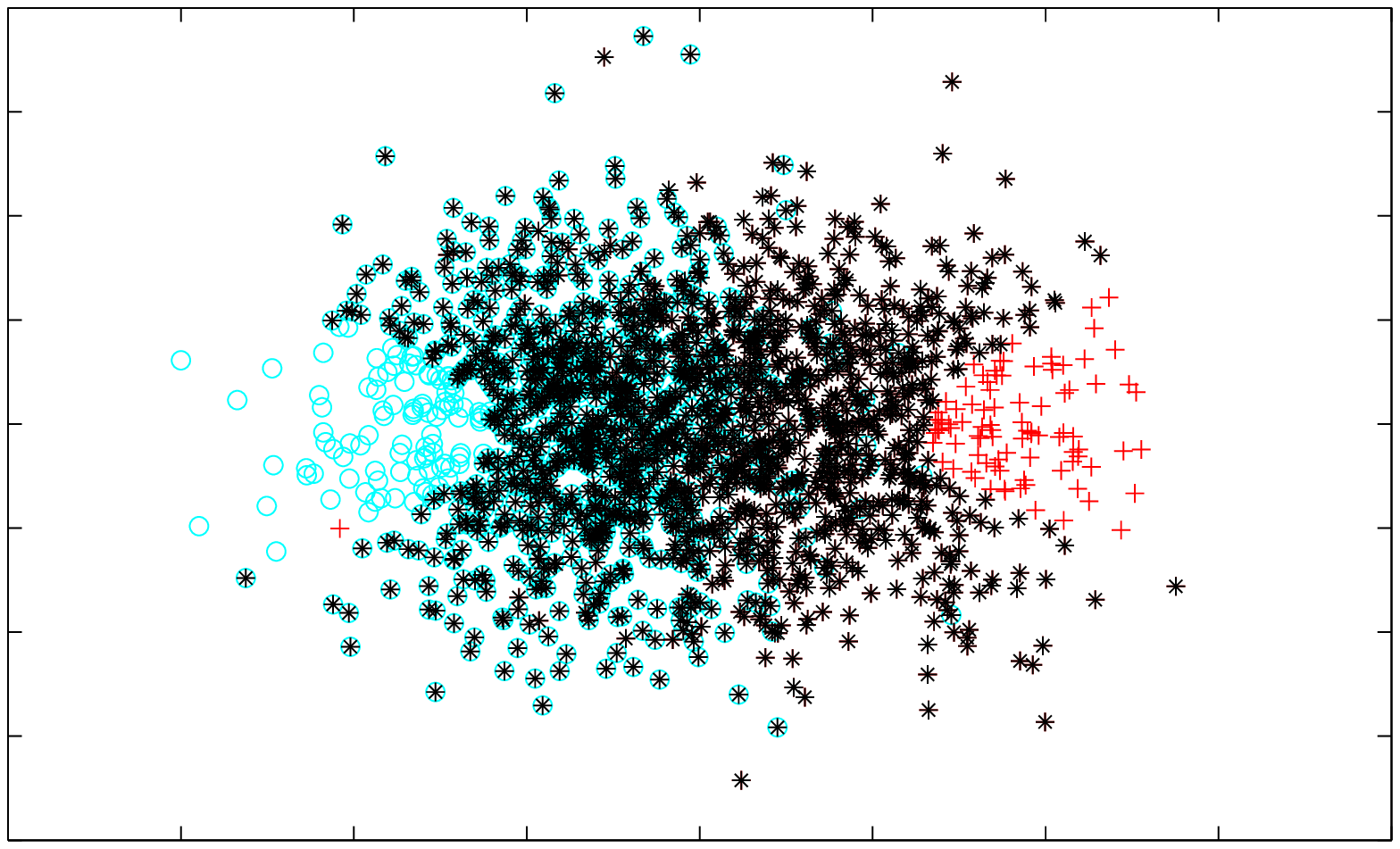}
\end{center}
\caption{Influence of the $\alpha$ parameter on selection of kernels for support features defined by vectors shown in the middle (here $\delta=0$). From top down: $\alpha=0.005, 0.05, 0.1$.}
\label{fig:alpha}
\end{figure}

\begin{figure}[!htb]
\begin{center}
\includegraphics[width=3.35in]{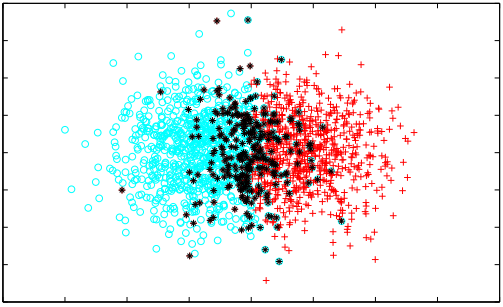}
\includegraphics[width=3.35in]{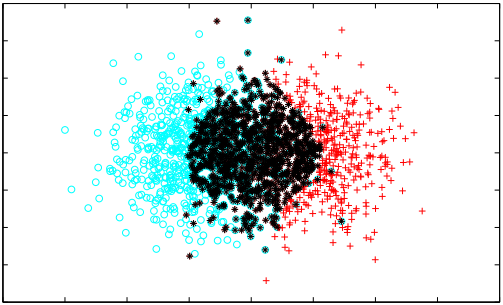}
\includegraphics[width=3.35in]{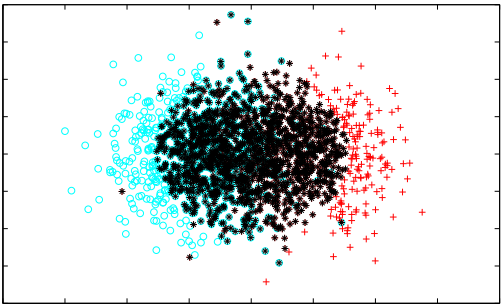}
\end{center}
\caption{Influence of the $\delta$ parameter on selection of kernels for support features defined by vectors shown in the middle (here $\alpha=0$). From top down: $\delta=0.5, 0.6$, and $0.7$.}
\label{fig:delta}
\end{figure}

\begin{figure}[!htb]
\begin{center}
\includegraphics[width=3.35in]{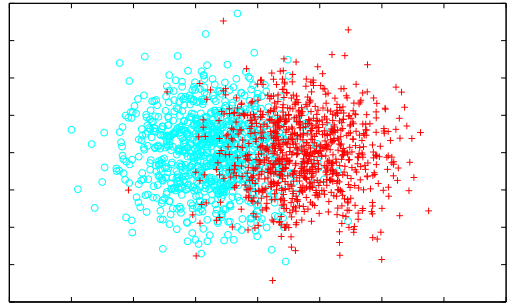}
\includegraphics[width=3.35in]{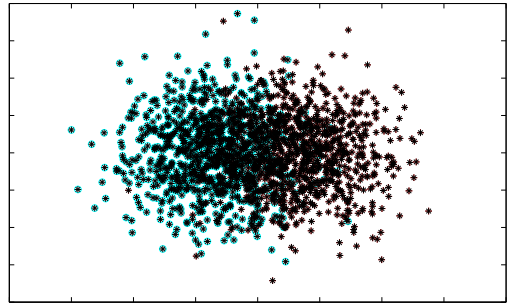}
\end{center}
\caption{Wrong selection of parameters leaves too few or too many kernel features. }
\label{fig:bad_parameters}
\end{figure}

\section{Illustrative examples}

The usefulness of new support feature has been tested on several benchmark  datasets, selected to cover different types of problems and to compare solutions with SVMs based on Gaussian kernels (on these datasets results with polynomial, Minkovsky and sigmoidal kernels have not been better), as well as other classifiers. 
Seven datasets have been downloaded from the UCI Machine Learning Repository \cite{UCIrep}. These datasets are standard examples of benchmark type and are used here to enable comparison of different learning methods. Missing feature values (if any) have been replaced by the mean values for a given class. A leukemia microarray gene expression data from \cite{Golub99} is an example of high-dimensional small sample problem. Leukemia has 7129 dimensions and it would be quite easy to get perfect results with such a large space, therefore only 100 best features from a simple Fischer Discriminant Analysis (FDA) ranking index have been used \cite{Duchfilter06}. 
In addition 8-bit parity dataset have been selected because it is very difficult to analyze correctly by standard Support Vector Machines or other machine learning algorithms. 
A summary of all datasets used is presented in Tab. \ref{tab:1}. 

Short description of the datasets used:
\begin{enumerate}
\item \textbf{Appendicitis} includes only 106 vectors, 8 attributes, two classes (85 acute and 21 other cases).
\item \textbf{Australian} has 690 cases of credit card applications, all 15 attribute names and values are changed to protect confidentiality of the data.
\item \textbf{Cleveland Heart} disease dataset with 303 samples, each described by 13 attributes, 150 cases labeled as ``absence", and 120 as ``presence" of heart disease. 
\item \textbf{Diabetes} dataset (also known as ``Pima Indian diabetes") contains 768 cases, with 500 negative, and 268 positive test results for diabetes. Each sample is described by 8 attributes. All patients were females at least 21 years old of Pima Indian heritage.
\item \textbf{Hepatitis} has 155 samples (32 from class 'die' and 123 from class 'live') characterized by 19 attributes, with many missing values.
\item \textbf{Ionosphere} has 351 data records, with 224 patterns in Class 1 and 126 in Class 2 (different types of radar signals reflected from ionosphere). First feature is binary, second is allways zero, the remaining 32 are continuous. 
\item \textbf{Leukemia} microarray gene expressions for two types of leukemia (ALL and AML), with a total of 47 ALL and 25 AML samples measured with 7129 probes. Evaluations of this data is based here on pre-selected 100 best features, done by simple feature ranking using FDA index.
\item \textbf{Parity8} 8-bit parity dataset, with 8 binary features and 256 vectors.
\item \textbf{Sonar} dataset contains signals obtained from a variety of different aspect angles, spanning 90 degrees for the cylinder (111 cases) and 180 degrees for the rock (97 cases). Each of 208 patterns is a set of 60 attributes.
\end{enumerate}

\begin{table}[!ht]
\caption{Summary of datasets used for tests} 
\begin{center}
\begin{tabular}{|@{\hspace{1.0mm}}c@{\hspace{1.0mm}}|@{\hspace{1.0mm}}c@{\hspace{1.0mm}}|@{\hspace{1.0mm}}c@{\hspace{1.0mm}}|@{\hspace{1.0mm}}c@{\hspace{1.0mm}}|@{\hspace{1.0mm}}c@{\hspace{1.0mm}}|}
\hline
Title & \#Features & \#Samples & \multicolumn{2}{|c|}{\#Samples per class}\\
\hline
Appendicitis	&	8	& 106	& 85 $C_1$	& 21 $C_2$ \\ \hline
Australian	&	15	& 690	& 307 positive	& 383 negative \\ \hline
Diabetes	&	8	& 768	& 500 negative	& 268 positive \\ \hline
Heart 		& 13		& 303 	& 160 absence 	& 137 presence \\ \hline
Hepatitis 	& 19 		& 155 	& 32 $C_1$	& 123 $C_2$ \\ \hline
Ionosphere 	& 34 		& 351 	& 224 $C_1$	& 126 $C_2$ \\ \hline
Leukemia  	& 100 		& 72 	& 47 ALL 	& 25 AML \\ \hline
Parity8  	& 8  		& 256 	& 128 even  	& 128 odd \\ \hline
Sonar  		& 60  		& 208 	& 111 metal 	& 97 rock \\
\hline
\end{tabular}
\label{tab:1}
\end{center}
\end{table}

\begin{table}[h]
\caption{Standard classifiers used in this paper}
\begin{center}
\begin{tabular}{|c|c|}
\hline
Classifier					& Short name \\ \hline
k-Nearest Neighbors				& kNN \\ \hline
Separability Split Value Tree \cite{SSV00}	& SSV \\ \hline
Support Vector Machines with Linear Kernel	& SVML \\\hline
Support Vector Machines with Gaussian Kernel	& SVMG \\ \hline
\end{tabular}
\label{tab:2}
\end{center}
\end{table}

To compare SFM with four popular classification methods (see Table \ref{tab:2}) 10-fold crossvalidation test results have been collected in Tables \ref{tab:svmsfm}-\ref{tab:SSVall}, with accuracies and standard deviations given for each dataset.
For the kNN classifier the number of nearest neighbors has been automatically selected from the $1-20$ range using crossvalidation estimation. The SVM parameters ($C$ and $\sigma$ for Gaussian kernels) have been fully optimized on the original data in an automatic way using crossvalidation estimations. 
Support features and all parameters have always been optimized within crossvalidation on the training partition only to be sure that no information about the whole data has been used at any stage. 
All calculations for standard classification methods have been performed using the Ghostminer package developed in our group \cite{GM}.

To check the influence of different types of support features all combinations have been investigated. Let's call the original features X, the kernel features K, the unrestricted linear projections Z, and the restricted (clustered) projections H. Then the following 15 feature spaces based on combinations of different type of features may be investigated: X, K, Z, H, K+Z, K+H, Z+H, K+Z+H, X+K, X+Z, X+H, X+K+Z, X+K+H, X+Z+H, X+K+Z+H. Unfortunately for all the classifiers used here this will make a very big table. Therefore only partial presentation of results is done below. 
First in Tab. \ref{tab:svmsfm} results of optimized SVM with linear (SVML) and Gaussian kernels (SVMG) are compared with SFM with added kernel features only. 

\begin{table}
\caption{SVM vs SFM in the kernel space only}
\begin{center}
\begin{tabular}{|c|c|c|c|}
\hline
Dataset		&SVML 		& SVMG 		& SFM(K) 	\\  \hline
Appendicitis	&87.6$\pm$10.3	&86.7$\pm$9.4	&86.8$\pm$11.0 	\\ \hline
Australian	&85.5$\pm$4.3  	&85.6$\pm$6.4	&84.2$\pm$5.6	\\ \hline
Diabetes	&76.9$\pm$4.5	&76.2$\pm$6.1	&77.6$\pm$3.1  	\\ \hline
Heart		&82.5$\pm$6.4	&82.8$\pm$5.1	&81.2$\pm$5.2 	\\ \hline
Hepatitis	&82.7$\pm$9.8	&82.7$\pm$8.4	&82.7$\pm$6.6 	\\ \hline
Ionosphere	&89.5$\pm$3.8	&94.6$\pm$4.4	&94.6$\pm$4.5 	\\ \hline
Leukemia	&98.6$\pm$4.5	&84.6$\pm$12.1	&87.5$\pm$8.1 	\\ \hline
Sonar		&75.5$\pm$6.9	&86.6$\pm$5.8	&88.0$\pm$6.4 	\\ \hline
Parity8		&33.4$\pm$5.9	&12.1$\pm$5.9	&11$\pm$4.3 	\\ \hline
\end{tabular}
\label{tab:svmsfm}
\end{center}
\end{table}

For ionosphere and sonar there is a big advantage in using the kernel space instead of the original features space and this is reflected also in the SFM(K) results. For Leukemia simple linear model works better as the number of patterns is very small. For parity all local neighborhoods contain only vectors from the wrong class so only if dispersions of Gaussian kernels are very large good solution is found (our automatic optimizer did not go that far). This examples shows two things: first, sometimes kernel features are less useful than the original features (and as we shall see below, projected features), and second, the differences between SVMG and SFM(K) are well within variance, so explicit representation in the kernel space gives equivalent solution. 

In fact best results have never been achieved in the kernel space only for any data and with any classifier we have tried (Tab. \ref{tab:sfmall}). This casts some doubt on the optimality of single kernel-based approaches. Also adding original inputs X have never been useful, therefore we shall not present these results here. Taking the SFM(K) results as the reference in Tab. \ref{tab:sfmall} influence of features space extensions on accuracy has been collected. Adding various types of support features leads to significant improvements, but for different data different types of feature seem to be important. In case of the Appendicitis the restricted projections lead to a significant improvement on 3\% with some reduction in variance. H features also increase accuracy of Heart on 3.6\% and on Hepatitis on 1.2\%. The most dramatic change is on the Parity data, where restricted projections allow to solve the problem almost perfectly (the reason why some errors are left is due to the fact that only clusters with at least 10 vectors are included as H features, this should be decreased to at most 8).
For Australian Credit and Leukemia the improvement was relatively small (about 2\%), and thus statistically not significant, therefore these datasets have been omitted in Table \ref{tab:sfmall}. 
Results for Ionosphere improve when kernel features are added and Sonar shows 3.9\% improvement for all types of features combined. 

\begin{table}
\caption{SFM in various spaces, see text for description.}
\begin{center}
\begin{tabular}{|@{\hspace{1.0mm}}c@{\hspace{1.0mm}}|@{\hspace{1.0mm}}c@{\hspace{1.0mm}}|@{\hspace{1.0mm}}c@{\hspace{1.0mm}}|@{\hspace{1.0mm}}c@{\hspace{1.0mm}}|@{\hspace{1.0mm}}c@{\hspace{1.0mm}}|@{\hspace{1.0mm}}c@{\hspace{1.0mm}}|}
\hline
Dataset		&  K		&	H	& 	K+H   &  Z+H	     & K+H+Z	\\  \hline
Appendicitis		&86.8$\pm$11 	&89.8$\pm$7.9  &89.8$\pm$7.9  &89.8$\pm$7.9  &89.8$\pm$7.9\\ \hline
Diabetes	&77.6$\pm$3.1  	&76.7$\pm$4.3  &79.7$\pm$4.3  &79.2$\pm$4.5  &77.9$\pm$3.3\\ \hline
Heart		&81.2$\pm$5.2 	&84.8$\pm$5.1  &80.6$\pm$6.8  &83.8$\pm$6.6  &78.9$\pm$6.7\\ \hline
Hepatitis	&82.7$\pm$6.6 	&83.9$\pm$5.3  &83.9$\pm$5.3  &83.9$\pm$5.3  &83.9$\pm$5.3\\ \hline
Ionosphere		&94.6$\pm$4.5 	&93.1$\pm$6.8  &94.6$\pm$4.5  &93.0$\pm$3.4  &94.6$\pm$4.5\\ \hline
Sonar		&83.6$\pm$12.6 	&66.8$\pm$9.2  &82.3$\pm$5.4  &73.1$\pm$11  &87.5$\pm$7.6\\ \hline
Parity8		&11$\pm$4.3 	&99.2$\pm$1.6  &97.6$\pm$2.0  &99.2$\pm$2.5  &96.5$\pm$3.4\\ \hline
\end{tabular}
\label{tab:sfmall}
\end{center}
\end{table}

Similar analysis may be performed for other methods in various spaces. 
The nearest neighbor algorithm (Table  \ref{tab:kNNall}) shows significant improvements, for example 8\% on the ionosphere in K+H space. 
Finally the SSV decision tree (Table  \ref{tab:SSVall}) in the K+H+Z space has improved a lot on data with continuous features, 
from 88 to 93.7\% on the ionosphere.

\begin{table}
\caption{kNN in various spaces}
\begin{center}
\begin{tabular}{|@{\hspace{1.0mm}}c@{\hspace{1.0mm}}|@{\hspace{1.0mm}}c@{\hspace{1.0mm}}|@{\hspace{1.0mm}}c@{\hspace{1.0mm}}|@{\hspace{1.0mm}}c@{\hspace{1.0mm}}|@{\hspace{1.0mm}}c@{\hspace{1.0mm}}|@{\hspace{1.0mm}}c@{\hspace{1.0mm}}|}
\hline
Dataset		&  X		&  H		& K+H         &  Z+H    & K+H+Z 	\\  \hline
Appendicitis		&86.7$\pm$6.6  &79.9$\pm$12  &81.1$\pm$5.8  &80.2$\pm$10.4  &83.8$\pm$9.5\\ \hline
Diabetes	&75.5$\pm$5.7  &76.7$\pm$4.3  &73.6$\pm$3.8  &76.8$\pm$4.6  &71.5$\pm$3.5\\ \hline
Heart		&82.2$\pm$7.3  &85.5$\pm$5.8  &82.9$\pm$8.8  &84.5$\pm$7.2  &82.8$\pm$8.2\\ \hline
Hepatitis	&83.3$\pm$7.6  &82.6$\pm$10.1  &83.0$\pm$11  &82.7$\pm$6.7  &83.4$\pm$8.0\\ \hline
Ionosphere		&86.3$\pm$4.4  &90.0$\pm$8.5  &94.6$\pm$4.5  &92.3$\pm$3.6  &94.6$\pm$4.7\\ \hline
Sonar		&86.5$\pm$4.5  &82.0$\pm$7.2  &82.5$\pm$8.4  &82.1$\pm$6.8  &84.9$\pm$9.0\\ \hline
Parity8		&100$\pm$0     &99.2$\pm$1.6  &100$\pm$0     &98.4$\pm$2.8  &100$\pm$0\\ \hline
\end{tabular}
\label{tab:kNNall}
\end{center}
\end{table}

\begin{table}
\caption{SSV in various spaces}
\begin{center}
\begin{tabular}{|@{\hspace{1.0mm}}c@{\hspace{1.0mm}}|@{\hspace{1.0mm}}c@{\hspace{1.0mm}}|@{\hspace{1.0mm}}c@{\hspace{1.0mm}}|@{\hspace{1.0mm}}c@{\hspace{1.0mm}}|@{\hspace{1.0mm}}c@{\hspace{1.0mm}}|@{\hspace{1.0mm}}c@{\hspace{1.0mm}}|}
\hline
Dataset		&  X		&  H		& K+H    &  Z+H    & K+H+Z 	\\  \hline
Appendicitis		&83.2$\pm$11   &86.2$\pm$9.5  &83.2$\pm$9.4  &87.9$\pm$7.5  &84.1$\pm$9.7\\ \hline
Diabetes	&73.0$\pm$4.7  &76.3$\pm$4.2  &72.8$\pm$3.6  &75.8$\pm$3.2  &76.0$\pm$4.7\\ \hline
Heart		&76.2$\pm$6.4  &84.2$\pm$5.0  &81.3$\pm$7.6  &82.2$\pm$5.6  &83.8$\pm$5.6\\ \hline
Hepatitis	&75.6$\pm$8.5  &85.3$\pm$8.3  &85.3$\pm$8.3  &80.7$\pm$11.2   &80.7$\pm$11\\ \hline
Ionosphere		&88.0$\pm$3.5  &93.8$\pm$3.4  &87.4$\pm$6.2  &93.2$\pm$4.3  &93.7$\pm$4.0\\ \hline
Sonar		&72.1$\pm$5.8  &64.3$\pm$8.9  &64.3$\pm$8.9  &73.1$\pm$13.6  &74.0$\pm$7.3\\ \hline
Parity8		&49.2$\pm$1.0  &98.5$\pm$2.7  &97.6$\pm$2.8  &95.3$\pm$5.2  &98.8$\pm$1.8\\ \hline
\end{tabular}
\label{tab:SSVall}
\end{center}
\end{table}

Summarizing, for Pima Indian Diabetes the best reported result was 77.7\% (variance not given) obtained with the Logdisc method \cite{statlog94}. SFM in K+H space has reached 79.7$\pm$4.3\%. On the other hand Raymer et al. \cite{Bayes:Raymer03} obtained 64-73\% using hybrid Bayes classifier/evolutionary algorithm optimizing feature subsets and kNN weights. 
For Cleveland Heart data SFM in H space gives 84.8$\pm$5.1\%, a relatively modest 2\% improvement over SVM. 
kNN reaches slightly higher 
85.5$\pm$5.8\% in the H space.

SFM has also achieved best results for the two problems with continuous features. 
On Sonar combination of all features leads to the SFM accuracy 87.5$\pm$7.6\%, showing the power of support features. Best MLP neural network results reported by Gorman and Sejnowski \cite{Sonar:Gorman88} are 84.7$\pm$5.7\%.  
Ionosphere also yielded good improvements in K+H feature space for all methods, with SFM results 94.6$\pm$4.5\%. For comparison, Raymer et al. \cite{Bayes:Raymer03} report 87-92.3\%. 

Australian Credit problem is also very popular \cite{UCIrep}, but it is usually approached in a wrong way. A single binary feature 
gives 85.5\% and it is easy to overlook creating more complex models \cite{duchieee04}. Here only SSV decision tree find slightly more accurate solution, but the improvement of 2.3\% in Z+H space may not be worth additional complexity. 

High-dimensional parity problem is very difficult for most classification methods. Many papers have been published on special neural models for parity functions, and the reason is quite obvious. Linear separation cannot be easily achieved because this is a $k$-separable problem that should be separated into $n+1$ intervals for $n$ bits \cite{DuchKsep06,GrochowskiD07}. This is a very interesting example showing that SFM solves quite easily difficult problems in almost perfect way even when most standard classifiers fails. Although kNN may also work perfectly well it requires $k>2n$ for $n$-bit parity to overcome the influence of the nearest neighbors, and will fail or less regular Boolean functions.  

\section{Discussion and conclusions}

Support Feature Machine algorithm introduced in this paper if focused on generation of new features rather than improvement in optimization and classification algorithms. A fruitful question is: what is the limit of accuracy for a given dataset that can be achieved in a given feature space? Progress in the recent years in classification and approximation methods allows us to be close to this limit in most cases, but, as the results obtained in this paper suggest, there is still ample room for improvement in generation of new features. For some data kernel-based features are important, for other projections and restricted projections discover more interesting aspects. Expanded feature space seems to benefit not only linear discriminators, but also nearest neighbor and decision tree methods much more than improvements of their algorithms. Recently more sophisticated ways of creating new features have also been introduced \cite{ULM09,Meta:Maszczyk10}, deriving them from various data models.  

SFM requires generation of new features, a process that is computationally efficient, followed by the selection of potentially relevant ones and used by any linear discrimination technique. Many variants of basic SFM algorithm are possible and the implementation reported here, although very successful, providing several results significantly better than others found in the literature, certainly is far from optimal. The goal was to fix all internal parameters at reasonable values, as it is done in SVM, where  also a number of parameters related to the solver are fixed. Better was to generate and select features will lead to more information extracted from data, and easier classification. For example, only binary H features  based on pure clusters have been considered, although soft windows may generate more interesting views on the data. More sophisticated thresholds for relevance of new features, weights proportional to the size of the clusters in restricted projections, or dynamic resolution based on distances for kernel features may be introduced.  
Mixing different kernels and using different types of features gives much more flexibility. Moreover, it is rather straightforward to introduce multiresolution in the SFM algorithm, for example using different dispersion $\beta$ for every $\vec{H}_j$. 
Kernel-based learning \cite{Scholkopf01} implicitly projects data into high-dimensional spaces, creating there flat decision borders end facilitating separability. The learning process is greatly simplified by changing the goal of learning to easier target and handling the remaining nonlinearities with well defined structure \cite{Duchfound07}. 
Adding support features facilitates also knowledge discovery. Instead of hiding information in kernels and sophisticated optimization techniques features based on kernels and projection techniques make this explicit. Intermediate representations are very important. Finding interesting views on the data, or constructing interesting information filters, is very important because combination of the transformation-based systems should bring us significantly closer to practical applications that automatically create the best data models for any data. 

It is also interesting to comment on neurobiological plausibility of the SFM approach. 
In \cite{CSKernels:Jakel09} authors argue that kernel methods are relevant for category learning in biological systems. In standard formulations of SVMs it is not quite obvious. However, the SFM algorithm may be presented in a network form, with the first hidden layer based on combination of kernels, projections, and localized projections. This corresponds to various functions of microcircuits that are present in cortical minicolumns. In effect this layer approximates liquid state machine \cite{Maass02}, while the output layer is a simple perceptron that reads off this information. With great diversity of microcircuits a lot of information is generated, and relevant chunks are used as features by simple Hebbian learning of weights in the output layer. In such model plasticity of the basic feature detectors receiving the incoming signals may be quite low, yet fast correlation-based learning is still possible.

\bibliographystyle{splncs}
\bibliography{CIpapers,KIS}

\end{document}